\documentclass[journal]{IEEEtran}

\usepackage{xspace}
\usepackage{balance}
\usepackage{graphicx}
\usepackage[hyphens]{url}
\usepackage{epstopdf}
\usepackage{microtype}
\usepackage{booktabs}
\usepackage[table,xcdraw]{xcolor}
\usepackage{caption}
\usepackage{subcaption}
\usepackage{amsmath}
\usepackage{enumitem}
\usepackage{listings}
\usepackage{color}
\usepackage{hyperref}
\usepackage{multirow}
\usepackage[numbers]{natbib} 

\usepackage{tabularx}
\newcolumntype{T}{>{\centering\arraybackslash}X}

\definecolor{mygreen}{rgb}{0,0.6,0}
\definecolor{mygray}{rgb}{0.5,0.5,0.5}
\definecolor{mymauve}{rgb}{0.58,0,0.82}

\lstdefinestyle{promptStyle}{
  backgroundcolor=\color{lightgray},
  basicstyle=\ttfamily\small,
  breaklines=true,
  frame=single
}

\lstdefinestyle{responseStyle}{
  backgroundcolor=\color{white},
  basicstyle=\ttfamily\small,
  breaklines=true,
  frame=single
}

\newcommand{\fakepar}[1]{\smallbreak\noindent{}}

\usepackage{manyfoot}
\DeclareNewFootnote{A}
\DeclareNewFootnote{B}

\let\footnote\footnoteA
\iftrue
    
\else
    \newcommand{\mb}[1]{}
\fi

\begin{document}

\title{Leveraging Large Language Models for DRL-Based Anti-Jamming Strategies in Zero Touch Networks}

\author{
Abubakar~S. Ali,~\IEEEmembership{Member,~IEEE,}
        Dimitrios~Michael~Manias,~\IEEEmembership{Member,~IEEE,}
        Abdallah~Shami,~\IEEEmembership{Senior~Member,~IEEE,}
        and Sami~Muhaidat,~\IEEEmembership{Senior~Member,~IEEE}
				\thanks{A.S. Ali is with the Department of Electrical and Computer Engineering, Khalifa University, Abu Dhabi 127788, UAE, (e-mails: 100052954@ku.ac.ae).}
				\thanks{D. M. Manias and A. Shami are with the Department of Electrical and Computer Engineering, Western University, Canada (e-mail: {dmanias3,abdallah.shami}@uwo.ca).}
		        \thanks{S. Muhaidat is with the KU Center for Cyber-Physical Systems, Department of Electrical Engineering and Computer Science, Khalifa University, Abu Dhabi 127788, UAE, and with the Department of Systems and Computer Engineering, Carleton University, Ottawa, ON K1S 5B6, Canada, (e-mail: muhaidat@ieee.org).}
		\vspace{-0.7cm}		}

\markboth{IEEE Journal,~Vol.~xx, No.~x, Month~2020}%
{Shell \MakeLowercase{\textit{et al.}}: Bare Demo of IEEEtran.cls for IEEE Journals}

\maketitle
    
\begin{abstract}
As the dawn of sixth-generation (6G) networking approaches, it promises unprecedented advancements in communication and automation. Among the leading innovations of 6G is the concept of Zero Touch Networks (ZTNs), aiming to achieve fully automated, self-optimizing networks with minimal human intervention. Despite the advantages ZTNs offer in terms of efficiency and scalability, challenges surrounding transparency, adaptability, and human trust remain prevalent. Concurrently, the advent of Large Language Models (LLMs) presents an opportunity to elevate the ZTN framework by bridging the gap between automated processes and human-centric interfaces. This paper explores the integration of LLMs into ZTNs, highlighting their potential to enhance network transparency and improve user interactions. Through a comprehensive case study on deep reinforcement learning (DRL)-based anti-jamming technique, we demonstrate how LLMs can distill intricate network operations into intuitive, human-readable reports. Additionally, we address the technical and ethical intricacies of melding LLMs with ZTNs, with an emphasis on data privacy, transparency, and bias reduction. Looking ahead, we identify emerging research avenues at the nexus of LLMs and ZTNs, advocating for sustained innovation and interdisciplinary synergy in the domain of automated networks.
\end{abstract}



\section{Introduction}
\label{sec:introduction}

The evolution of modern networks has propelled us from the rudimentary beginnings of communication to the intricate web of interconnected systems that define our digital age. With the advent of sixth-generation (6G) networking, we stand on the brink of a paradigm shift, poised to unlock unprecedented advancements in communication and automation \cite{chergui2023toward}. Central to this evolution is the visionary concept of Zero Touch Networks (ZTNs), ushering in an era of autonomous, self-optimizing networks with minimal human intervention \cite{alwis20236G, ericsson2023zero}. As the possibilities of 6G expand, seamless integration of these cutting-edge paradigms with the tools and capabilities that enable their realization becomes paramount \cite{chergui2023toward}.

Critical to the empowerment of ZTNs is the realm of reinforcement learning (RL). Techniques like deep reinforcement learning (DRL) have been harnessed to forge adaptive strategies that counter jamming attacks, enhancing network robustness and reliability. However, the intrinsic black-box nature of DRL raises interpretability challenges. As DRL agents extract intricate policies from data, understanding the decision-making process and rationale behind their actions becomes complex \cite{du2023guiding}. This opacity is particularly concerning in safety-critical applications like wireless communication systems, where comprehending agent behavior is imperative \cite{du2023power}. Researchers and practitioners are diligently exploring methods to enhance the interpretability of DRL models. Mechanisms such as attention mechanisms, visualization tools \cite{chen2021deep}, and explainable AI approaches \cite{rjoub2022explainable} have emerged, aimed at shedding light on the pathways through which DRL agents arrive at decisions. Striking a balance between DRL's adaptability and efficacy and the need for transparency remains an ongoing research frontier.

Concurrently, the ascent of Large Language Models (LLMs) has revolutionized artificial intelligence, endowing machines with unprecedented language understanding, generation, and manipulation capabilities. Trained on extensive textual corpora, these models excel at extracting insights, crafting narratives, and emulating human interactions \cite{bariah2023large}. The convergence of ZTNs and LLMs bridges the gap between machine-driven automation and human-centric interfaces \cite{chergui2023toward}. A substantial body of research has emerged at the intersection of network automation, language modeling, and reinforcement learning. While ZTNs focus on self-optimizing network architectures, LLMs introduce novel approaches to natural language understanding and generation \cite{bowman2023things}.

The fusion of Large Language Models (LLMs) with diverse domains underscores their ability to bridge the gap between intricate technical processes and human understanding. Noteworthy studies have showcased LLMs' prowess. In \cite{ferrag2023revolutionizing}, the authors introduce SecurityLLM, an adept cyber threat detection model with components SecurityBERT and FalconLLM, proficiently identifying 14 attack types via text analysis. This pioneering effort outperforms conventional methods, achieving 98\% accuracy on evaluated cybersecurity data. \cite{charalambous2023new} presents an innovative approach amalgamating LLMs and formal verification, successfully detecting and repairing software vulnerabilities in C programs with an 80\% success rate. Furthermore, \citet{ferrag2023securefalcon} introduce SecureFalcon, refining FalconLLM~\cite{falcon40b} to identify vulnerabilities in C code with 94\% accuracy, showcasing the potential for enhancing software quality and security. This paper, representing the first of its kind in the literature, delves into the symbiotic relationship between LLMs and ZTNs, revealing how their integration enhances network transparency, interaction, and decision-making, empowering stakeholders to influence autonomous network behavior.

The paper is organized as follows: Section~\ref{sec:ZTNs_overview} provides an overview of zero-touch networks. Section~\ref{sec:LLMs_overview} presents an overview of large language models' capabilities and relevance. Section~\ref{sec:anti-jamming} offers an overview of DRL-based anti-jamming, which is studied as a prominent use case in this paper. In Section~\ref{sec:integration}, we delve into the integration of LLMs in ZTNs, specifically through a comprehensive understanding of the DRL-based anti-jamming use case. Section~\ref{sec:considerations} highlights the technical and ethical challenges of employing LLMs in ZTNs. Moving forward, Section~\ref{sec:directions} explores the potential future directions in the realm of LLM-ZTN research. Finally, we conclude in Section~\ref{sec:conclusions}.

\section{Zero Touch Networks: Overview and Challenges}
\label{sec:ZTNs_overview}

\subsection{Overview of ZTNs}
In the modern era, network operators confront challenges due to the rapid progression of next-generation networks and an upsurge in connectivity demands~\cite{ericsson2023zero}. Zero-Touch Network Service Management (ZSM), as defined by ETSI, offers a solution, focusing on end-to-end network automation. It embodies four pivotal characteristics: self-configuration, self-monitoring, self-healing, and self-optimization~\cite{alwis20236G,chergui2023toward}.

For these characteristics to manifest, ZSM embraces several guiding principles. \textit{Modularity} ensures services are distinct and independent. \textit{Extensibility} allows the network to grow without hampering its existing operations, while \textit{Scalability} adapts to fluctuating user demands. The \textit{Model-Driven} structure uses domain-specific designs for effective service management~\cite{chergui2023toward}. \textit{Closed-Loop Management Automation} targets self-optimization, and \textit{Stateless Function Support} separates processing from storage needs. Principles like \textit{Resilience}, \textit{Management Domains}, and \textit{Service Composability} ensure network stability, organized resource handling, and integrated services, respectively. \textit{Intent-Based Interfaces} and \textit{Functional Abstraction} simplify operations, while \textit{Design for Automation} embeds automation at the core of every network component~\cite{alwis20236G}.

\subsection{Architecture of ZTNs}
The realization of the ZTNs through the application of the ZSM principles yields a set of architectural components that form the fundamental building blocks of these networks~\cite{ericsson2023zero,alwis20236G}. The first building block is the \textit{Management Services}. These services are at the most granular architectural level and offer services for consumption. Multiple entities can consume these services and can \subsection{ZTN Architectural Components}

Management services, fundamental to ZTNs, can amalgamate to create composite services~\cite{chergui2023toward}. Above these, the \textit{Management Function} serves as both a consumer and producer of such services, adapting its role based on its operational behavior.

The \textit{Management Domain} delineates areas with distinct administrative rules, structured around operational, functional, and deployment constraints~\cite{alwis20236G}. They house management functions and services, orchestrating resources ranging from physical to cloud-based. These domains can be partitioned further, with services classified as internal or external based on their accessibility.

Distinctively, the \textit{End-to-End Service Management Domain} oversees customer-centric services, diverging from traditional domains by not directly handling resources.

The ZTN's backbone, \textit{Integration Fabric}, ensures cohesive interaction among management functions. It provides essential services like discovery and registration, often represented as a collective or individual management function~\cite{chergui2023toward}.

Conclusively, \textit{Data Services} play a pivotal role, bridging data communication across management domains and reinforcing the ZSM's core principle of segregating data storage from processing~\cite{alwis20236G}.

\subsection{Challenges of ZTNs}
\subsubsection{AI Maturity and Data Availability}
One of the greatest challenges hindering the development and adoption of ZTMs is the limitations of the current state-of-the-art machine learning and artificial intelligence methods~\cite{du2023guiding, bowman2023things}. Despite the significant progress the field has made in the past decade, the complexity of next-generation networks still poses a significant challenge to network operators in terms of modelling~\cite{du2023power}. Furthermore, the challenge of AI explainability must mature significantly before implementations of this magnitude and of this criticality can be safely realized. Additionally, there is a major lack in the availability of accessible and comprehensive network datasets capturing a wide range of network events and behaviours. As machine learning models are dependent on the quality of data used for training, this unavailability poses a significant barrier to the development of a good model~\cite{chen2021deep}.

\subsubsection{Model Drift}
A second challenge that has been extensively investigated in the past is the notion of model drift which appears when the training environment of a model is different from the environment in which it is deployed. Model drift can have several causes; however, its manifestation always results in a model’s performance degradation. There are two main types of model drift, data drift and concept drift. Data drift describes a situation in which the data used during model training is statistically different from the data seen during active deployment. Conversely, concept drift describes a situation where the relationship learned during training is different from the relationship observed during deployment. Several strategies have been proposed to combat model drift in dynamic network environments, including drift-resistant architectures and drift detection and adaptation methods~\cite{rjoub2022explainable}.

\subsubsection{Security}
As networks become increasingly more distributed and diverse, the potential area of attack for malicious agents increases exponentially. An area of major concern is the security of machine learning models during training and operation since these models form the foundation of the ZTN~\cite{ferrag2023revolutionizing}. If the training phase of machine learning models is tampered with through false data, the integrity of the resulting model is severely compromised and can lead to false insights and flawed decision-making. Furthermore, the tampering of deployed machine models, either through false observations or poisoned insights, can have devastating effects on the quality of the network. When considering the widespread integration of machine learning in future networks and its critical role in ZTNs, it is evident that compromised machine learning insights can lead to severe management and orchestration errors that can jeopardize the stability of the network and the safety of its information and users~\cite{ferrag2023securefalcon}.


\section{Large Language Models: Capabilities and Relevance}
\label{sec:LLMs_overview}

\subsection{Fundamentals of LLMs}
Traditionally, methods such as recurrent neural networks have been the de facto standard for NLP tasks~\cite{chen2021deep}. One of the reasons this family of models, in particular, gained significant popularity in recent years has been their ability to learn sequential relationships in data, something which distinguishes them as candidates for NLP tasks~\cite{du2023guiding}. As research in this field progressed, one key challenge emerged that would ultimately lead to an implementation barrier and the need to explore alternative solutions, the scalability problem. Despite their ability to learn sequential patterns, RNNs suffer from scalability concerns both in the model training and deployment phases~\cite{bowman2023things}. To this end, transformer networks emerged as the successor of RNNs for NLP tasks and have since taken off by displaying incredible potential with the introduction of LLMs such as ChatGPT and Google Bard~\cite{falcon40b}. Transformer neural networks are a revolutionary neural network architecture that leverages parallelization and the concept of attention to enable the efficient and scalable training of massive models with billions of tunable parameters with vast amounts of data~\cite{du2023power}.

\subsection{Natural Language Processing}
Today’s LLMs have incredible capabilities, potential, and limitations~\cite{falcon40b}. In terms of capabilities, state-of-the-art LLMs excel at all types of text generation, including next-word prediction, text summarization, as well as editing and adjusting based on a desired tone or set of requirements~\cite{du2023guiding}. Reaching beyond the level of simple text generation, LLMs also have the ability to generate code based on a description of required functionalities as well as solve math problems to a high degree~\cite{du2023power}. Despite these incredible feats, the current state-of-the-art has only scratched the surface of what LLMs can do since model sizes and training datasets are exponentially increasing with each new iteration~\cite{bowman2023things}. As models continue increasing their complexity and the data with which they are trained becomes more diverse, the capabilities of LLMs will continue to develop, and potential use cases that are currently inconceivable will be realized~\cite{bariah2023large}. Despite their incredible potential, there are still causes of concern with these models, mainly related to bias and ethics. Depending on the nature of the data being used to train these models, LLMs will exhibit the implicit and explicit biases that are found within~\cite{du2023guiding}. This can be problematic as the model might produce undesirable or offensive content that could potentially be riddled with misinformation and disinformation. For this reason, extreme caution must be taken when training and deploying these models with constant monitoring and testing to ensure the desired results and behaviour are exhibited~\cite{bowman2023things}.

\subsection{LLMs in Networking}
LLMs present a lucrative opportunity for implementation in networking settings~\cite{du2023power, du2023guiding}. One of their potential implementations is to act as an interface between the network and the customer. In such a deployment, the customer would request a specific network setup or a change in their configured network, and the LLM would be tasked with interpreting the customer request and converting that into a set of commands that could be directly implemented in the network~\cite{ericsson2023zero}. Additionally, LLMs can be used to troubleshoot customer issues on a large scale, thereby speeding up customer support times~\cite{bariah2023large}. By leveraging LLMs to handle customer-facing tasks and operations, the speed at which requests are received and processed can be greatly reduced, thereby improving the system’s efficiency as well as the customer’s quality of experience~\cite{falcon40b}. Another opportunity for LLMs is the notion of explainability and transparency in the network. The LLM can be used to explain decisions taken by autonomous network agents or to describe network conditions to a network operator~\cite{du2023guiding}. This added level of interaction will lead to much more transparent networks that are easier to diagnose, troubleshoot, and manage.

\section{Deep Reinforcement Learning and Anti-Jamming}
\label{sec:anti-jamming}

Modern communication systems are susceptible to a range of security threats, and one of the prominent challenges is dealing with jamming attacks that can disrupt wireless communications \cite{Ali2022Deep}. DRL has emerged as a promising approach to tackle such anti-jamming problems by allowing intelligent agents to autonomously learn effective strategies to counteract jamming attempts. In this section, we delve into the concept of DRL-based anti-jamming, exploring how it leverages RL principles to enhance the robustness of wireless communication systems.

\subsection{Principles of DRL}
DRL combines Deep Learning (DL) techniques with RL principles to enable agents to learn optimal actions within an environment to achieve specific goals. At its core, DRL involves interacting with an environment, making a sequence of actions, and receiving feedback in the form of rewards. The agent learns to maximize the cumulative reward over time by iteratively adjusting its actions based on the received feedback. RL comprises several key elements, including:

\begin{itemize}
    \item \textbf{Agent}: The entity that interacts with the environment and learns to make decisions.
    \item \textbf{Environment}: The context in which the agent operates and learns. It provides feedback in response to the agent's actions.
    \item \textbf{State}: The representation of the current situation or condition of the environment.
    \item \textbf{Action}: Choices made by the agent to influence the environment.
    \item \textbf{Reward}: The feedback signal that indicates the desirability of an agent's action.
    \item \textbf{Policy}: A strategy or mapping that defines the agent's behavior by specifying actions in different states.
    \item \textbf{Value Function}: An estimate of the expected cumulative reward that an agent can obtain from a given state.
\end{itemize}

\subsection{DRL-based Anti-Jamming}
DRL has found applications in various domains, ranging from robotics and gaming to finance and healthcare. In the context of anti-jamming, DRL has shown significant promise in addressing the challenges posed by jamming attacks in wireless communication systems. The anti-jamming problem can be formulated as a Markov Decision Process (MDP), where the agent learns a policy to select appropriate communication channels in the presence of jamming.

\subsubsection{MDP Formulation}
In our anti-jamming scenario, the agent's state space comprises information about the received jamming power across different communication channels defined by a state vector $\mathbf{P_t}$. Specifically, $\mathbf{P_t} = [p_{t,1}, p_{t,2}, \cdots, p_{t,N_c}]$, with $p_{t,i}$ representing received power for frequency $i$ at time $t$. The state space size is $\left |\mathcal{S} \right | = N_c$. The action space corresponds to the choice of a channel for communication. The agent's objective is to select channels that are less susceptible to jamming, thereby maximizing communication throughput. This objective leads to rewards defined by considering whether the transmitter's selected frequency $f_{T,t}$ matches the jammer's frequency $f_{J,t}$. When $f_{T,t}$ and $f_{J,t}$ are different, the reward function is given by $\mathcal{U}(f_{T,t}) -\Gamma \delta (a_t\neq a_{t-1})$, where $\Gamma$ represents the cost incurred due to channel switching. Conversely, when both frequencies match, the reward is set to 0. Transitioning from state $s_t$ to $s_{t+1}$ upon reward $r_t$ follows transition probability $p(s_{t+1}|s_t, a_t)$. By interacting with the environment and observing the consequences of its actions, the DRL agent gradually learns a policy that adapts to the jamming dynamics. The agent can learn to switch channels strategically, taking into account the received jamming power, adjacent channel interference, and $\Gamma$.

\subsubsection{Agent Design}
Our approach employs a Double Deep Q-Network (DDQN) agent, an advanced variant of the standard Deep Q-Network (DQN), which mitigates Q-value overestimation by utilizing distinct networks for current and target Q-value estimation \cite{Ali2022Deep}. The DQN, a model-free, online, off-policy RL technique, leverages a value-based RL agent to train a Q-network that estimates future rewards. This selection is apt for our scenario due to the continuous observation space and discrete action space. The DQN agent employs a neural network as a function approximator, with $\theta_{Q}$ denoting its updateable weights. The Q-network comprises two hidden layers, each of 256 neurons, employing the ReLU activation function $f(x)=\textup{max}(0,x)$ \cite{Ali2022Deep}. The experience replay buffer $\mathcal{D}$ retains the agent's experiences as transition pairs, defined as $(s_{t},a_{t},r_{t},s_{t+1})$ at time-step $t$. To train, we utilize the stochastic gradient descent (SGD) algorithm \cite{Ali2022Deep} to update weights $\theta_{t}$ at each time-step $t$.

\section{Integrating LLMs into ZTNs}
\label{sec:integration}
Marrying LLMs with ZTNs ushers in a novel approach to network management, enabling complex automated processes to be conveyed in relatable human terms. This section provides a detailed exploration of the integration, emphasizing the pivotal role of the `falcon 7B` in generating insights from DRL-agent training.

\begin{figure}
    \centering
    \includegraphics[width=\columnwidth]{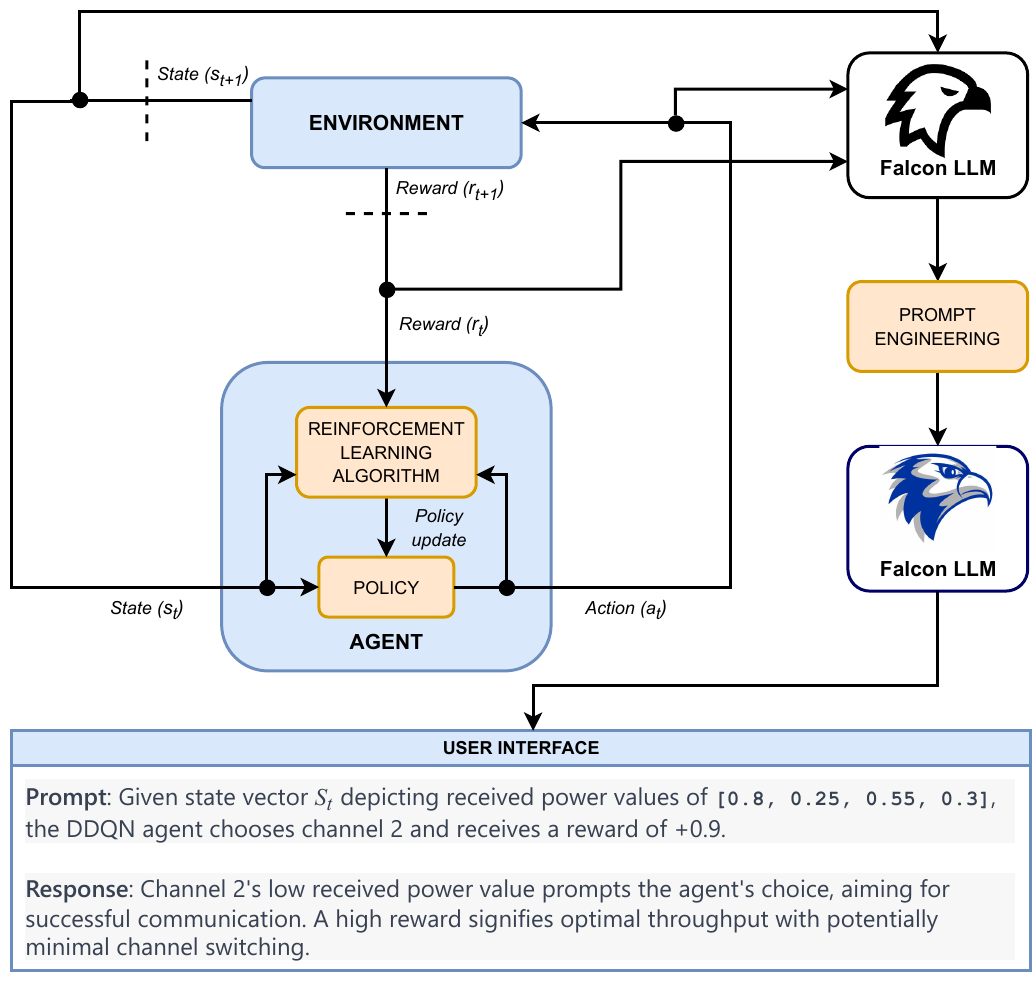}
    \caption{Integration schematics of the DRL-based anti-jamming technique combined with LLM interpretation.}
    \label{fig:arch}
\end{figure}

\subsection{Integration Mechanics}
\subsubsection{Data Handling and Processing}
ZTNs inherently manage a vast array of complex network data. Within this realm, the integration of the `falcon 7B` LLM amplifies the capacity for insightful data interpretation. Fig. \ref{fig:arch} demonstrates the schematic that elucidates how data from the DRL-agent training is ingested and processed. In our codebase,  (\url{https://huggingface.co/spaces/asataura/jam_shield_LLM_app}) we use prompt engineering in the `train` function within our `trainer` module to showcase this. Here data about rewards, rolling averages, and epsilon values are fed to the LLM, to facilitate a richer understanding and optimizing network decisions.

\subsubsection{LLM as an Interpreter}
At the heart of the LLM-ZTN synergy is the transformative capability of the `falcon 7B` to act as an interpreter. It interprets intricate DRL strategies and actions, like those in our anti-jamming approach, and converts them into clear, actionable insights. Using a tailored prompt template, complex numerical data, and DRL-agent states are translated into human-readable narratives, bridging the gap between raw data and actionable insights.

\subsubsection{Interactive Queries}
The inclusion of LLMs, particularly the `falcon 7B`, within the ZTN framework provides an interactive layer. Through the Streamlit UI, network administrators can actively engage with the system, seeking clear explanations for automated decisions. This interaction fosters trust, empowering operators with a deeper understanding, as evidenced in our training insights generation.

\subsubsection{Training and Feedback}
Beyond mere elucidation, the `falcon 7B` LLM offers evaluative feedback. Drawing from our DRL-based anti-jamming technique, the LLM assesses algorithmic performance, spotlighting both areas of success and potential refinements. This dual role of the LLM, both as an interpreter and evaluator, underscores its value in the ZTN framework. 

\begin{figure}
    \centering
    \includegraphics[width=\columnwidth]{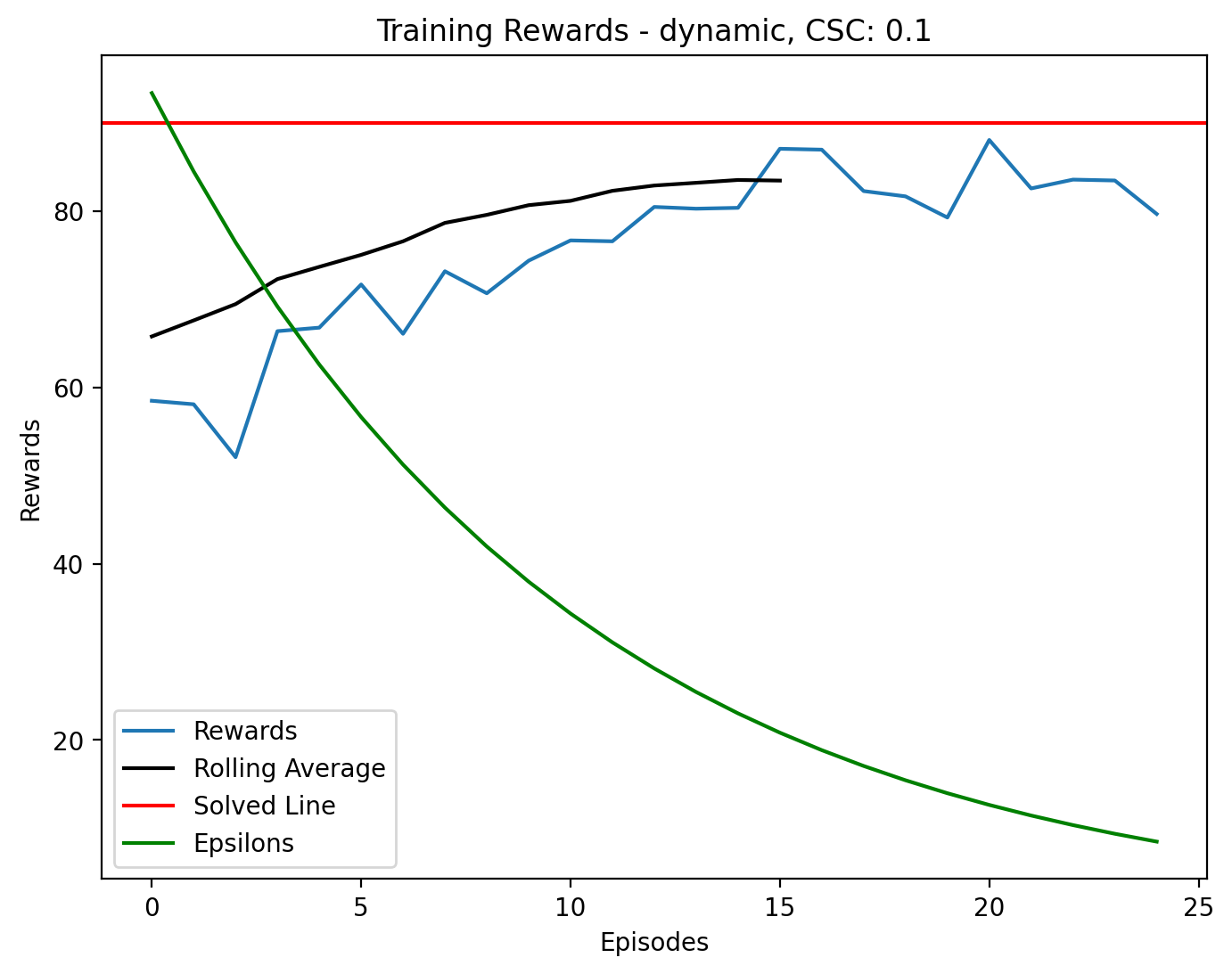}
    \caption{Training performance of the DRL-based anti-jamming agent.}
    \label{fig:performance}
\end{figure}

\begin{table}[h!]
    \centering
    \begin{tabular}{p{0.45\textwidth}}
        \toprule
        \textbf{Prompt Input} \\
        \midrule
        \begin{lstlisting}[style=promptStyle]
The graph represents training rewards over 25 episodes. The actual rewards range from 52.10 to 88.10 with an average of 75.50. The rolling average values range from 65.80 to 83.56 with an average of 77.27. The epsilon values decrease from 0.93 to 0.08 over the episodes. The solved threshold is set at 90.
        \end{lstlisting} \\
        \midrule
        \textbf{Falcon 7B Response} \\
        \midrule
        \begin{lstlisting}[style=responseStyle]
The graph represents the average training rewards over a set of episodes, with the range of rewards ranging from 52.10 to 88.10 and an average of 75.50. The rolling average values range from 65.80 to 83.56, with an average of 77.27. The epsilon values range from 0.08 to 0.93, with an average exploration rate of 0.36. The solved threshold is set at 90.00. This graph provides insights into the effectiveness of the algorithm in finding good solutions for the problem of deep reinforcement learning.
        \end{lstlisting} \\
        \bottomrule
    \end{tabular}
    \caption{Sample prompts and corresponding insights of the DRL training performance generated by Falcon 7B LLM.}
    \label{table:llm_insights_codeformat}
\end{table}

\subsection{Case Study: Anti-Jamming}
The potential of LLM integration into ZTNs is vividly demonstrated through our case study on DRL-based anti-jamming techniques. Here, the `falcon 7B` LLM plays an instrumental role in transforming raw training data used to generate a visualization depicted in Fig. \ref{fig:performance} into coherent reports. The engineered prompt and response of the `falcon 7B` LLM is demonstrated in Table~\ref{table:llm_insights_codeformat}. This use case not only exemplifies the technical prowess of the integration but also underscores its value in enhancing transparency and trust in automated networks.

Peering ahead, the melding of LLMs, particularly models like `falcon 7B`, with ZTNs signifies a promising fusion of automation with human-centric insights. Ensuring this integration is both seamless and meaningful will be pivotal in steering the future trajectory of automated networks.

\section{Technical and Ethical Considerations}
\label{sec:considerations}
The integration of LLMs into ZTNs, while promising, is riddled with both technical and ethical intricacies~\cite{du2023guiding, bariah2023large}. As LLMs interpret and elucidate automated network operations, they inherently access and process vast amounts of data~\cite{falcon40b}. This brings forth concerns about data privacy, decision accountability, and potential biases inherent in the models~\cite{bowman2023things}. The following sections delve deeper into these aspects, laying out the challenges and potential remedies.

\subsection{Technical Hurdles}
The amalgamation of LLMs with ZTNs poses several technical challenges~\cite{du2023power}. LLMs require substantial data for effective operation. Managing, storing, and efficiently retrieving this data, especially in real-time scenarios, is a significant hurdle~\cite{falcon40b}. For ZTNs to be truly responsive, LLMs need to interpret and respond in real-time. Achieving this speed, without compromising on the quality of insights, remains a challenge~\cite{chergui2023toward}. Furthermore, ZTNs are dynamic environments. LLMs must be adaptable to ever-changing network conditions, requiring continuous learning and updating mechanisms~\cite{ericsson2023zero}.

\subsection{Ethical Dimensions}
The integration of LLMs into automated systems introduces a spectrum of ethical concerns~\cite{du2023guiding, bowman2023things}. LLMs have access to vast datasets, which might include sensitive information. Ensuring that these models respect user privacy and don't inadvertently leak or misuse data is paramount~\cite{ferrag2023securefalcon, ferrag2023revolutionizing}. While LLMs aim to provide clarity on automated decisions, the inner workings of these models are intricate. Ensuring that they remain transparent and interpretable is crucial for maintaining trust~\cite{bowman2023things}. Additionally, LLMs, like all models, are trained on data. If this data contains biases, the model will inherit and potentially amplify them. Addressing these biases is not just a technical challenge but an ethical imperative~\cite{du2023guiding}.

\subsection{Bias and Fairness}
The ethical mandate of any AI system, including LLMs, is to ensure unbiased decision-making~\cite{bowman2023things}. However, biases in training data or model architecture can skew insights~\cite{du2023guiding}. Regular audits of the LLM's decisions can help identify biases in its recommendations or interpretations~\cite{bowman2023things}. LLMs should be designed to learn continuously, adjusting and rectifying biases from real-world feedback~\cite{du2023guiding}. Ensuring that the training data is representative and diverse can reduce inherent biases~\cite{bowman2023things}.

\subsection{Security Implications}
Integrating LLMs into ZTNs introduces new vectors for potential threats~\cite{ferrag2023securefalcon, ferrag2023revolutionizing}. With LLMs processing vast amounts of data, they become attractive targets for breaches. Robust security measures are essential to protect this data~\cite{ferrag2023revolutionizing}. Adversaries might attempt to tamper with the LLM to skew its decisions. Mechanisms to detect and counteract such tampering are crucial~\cite{ferrag2023securefalcon}. Ensuring that every decision or recommendation made by the LLM can be traced and justified is essential for both trust and security~\cite{charalambous2023new}.

\section{Future Directions in LLM-ZTN Research}
\label{sec:directions}
The fusion of LLMs and ZTNs represents a novel interdisciplinary frontier, with vast untapped potential. As both fields witness rapid advancements, their convergence promises to reshape the landscape of automated networks. This section explores the prospective evolution of LLMs and ZTNs and envisions the research avenues that lie at their intersection.

\subsection{Technological Prospects}
The domains of ZTN and LLM are poised for several groundbreaking advancements~\cite{du2023guiding, alwis20236G, chergui2023toward, falcon40b}. The future of ZTNs is in networks that can not only self-optimize but also self-heal and self-configure based on contextual needs~\cite{alwis20236G, chergui2023toward}. As LLMs grow in capacity, they will likely evolve to understand context better, make more nuanced interpretations, and even anticipate user queries~\cite{du2023guiding, falcon40b}. Future iterations might witness ZTNs that can directly communicate their learning needs to LLMs, fostering a seamless loop of learning and feedback~\cite{du2023guiding, du2023power}.

\subsection{Emerging Research Avenues}
The integration of LLMs and ZTNs opens a plethora of research possibilities~\cite{du2023guiding, bowman2023things, ferrag2023securefalcon, ferrag2023revolutionizing}. Researching methods to reduce the latency in LLM responses to facilitate real-time ZTN adjustments is a prime focus~\cite{du2023power, rjoub2022explainable}. There is also an emphasis on techniques that can detect and counteract biases in LLM interpretations, especially in the context of network decisions~\cite{bowman2023things}. Furthermore, ensuring that the communication between LLMs and ZTNs is secure, robust against adversarial attacks, and tamper-proof is of paramount importance~\cite{ferrag2023securefalcon, ferrag2023revolutionizing}.

\subsection{Interdisciplinary Collaborations}
The fusion of LLMs and ZTNs is inherently interdisciplinary, demanding expertise from diverse domains~\cite{du2023guiding, bowman2023things, chen2021deep, falcon40b}. The symbiosis between AI models (like LLMs) and networking techniques (like ZTNs) will be central to the evolution of automated networks~\cite{du2023guiding, du2023power, chen2021deep}. As AI models play a more significant role in decision-making, experts in ethics must guide their integration to ensure fairness, transparency, and accountability~\cite{bowman2023things}. Understanding how humans interact with and trust AI-driven networks will be pivotal, demanding a blend of psychology, design, and technical expertise~\cite{falcon40b}.

\section{Conclusions}
\label{sec:conclusions}

The rise of Zero Touch Networks (ZTNs) has revolutionized network automation. Yet, with their increasing complexity, there's a pressing need for clarity and transparency. Large Language Models (LLMs) offer a solution, especially evident in the context of demystifying DRL-based anti-jamming techniques. As intermediaries, LLMs translate complex network behaviors into digestible insights, bridging automation and human understanding. However, this integration is nascent and not without challenges. Both technical hurdles and ethical considerations demand attention. As we move forward, continuous research and interdisciplinary collaboration will be essential to fully realize the potential of LLMs in ZTNs. In essence, the fusion of LLMs with ZTNs holds immense promise, marking a pivotal step towards more transparent and human-centric automated networks.


\bibliographystyle{IEEEtranN}
\bibliography{references/bibliography}

%

\begin{IEEEbiography}[{\includegraphics[width=1in,height=1.25in, clip, keepaspectratio]{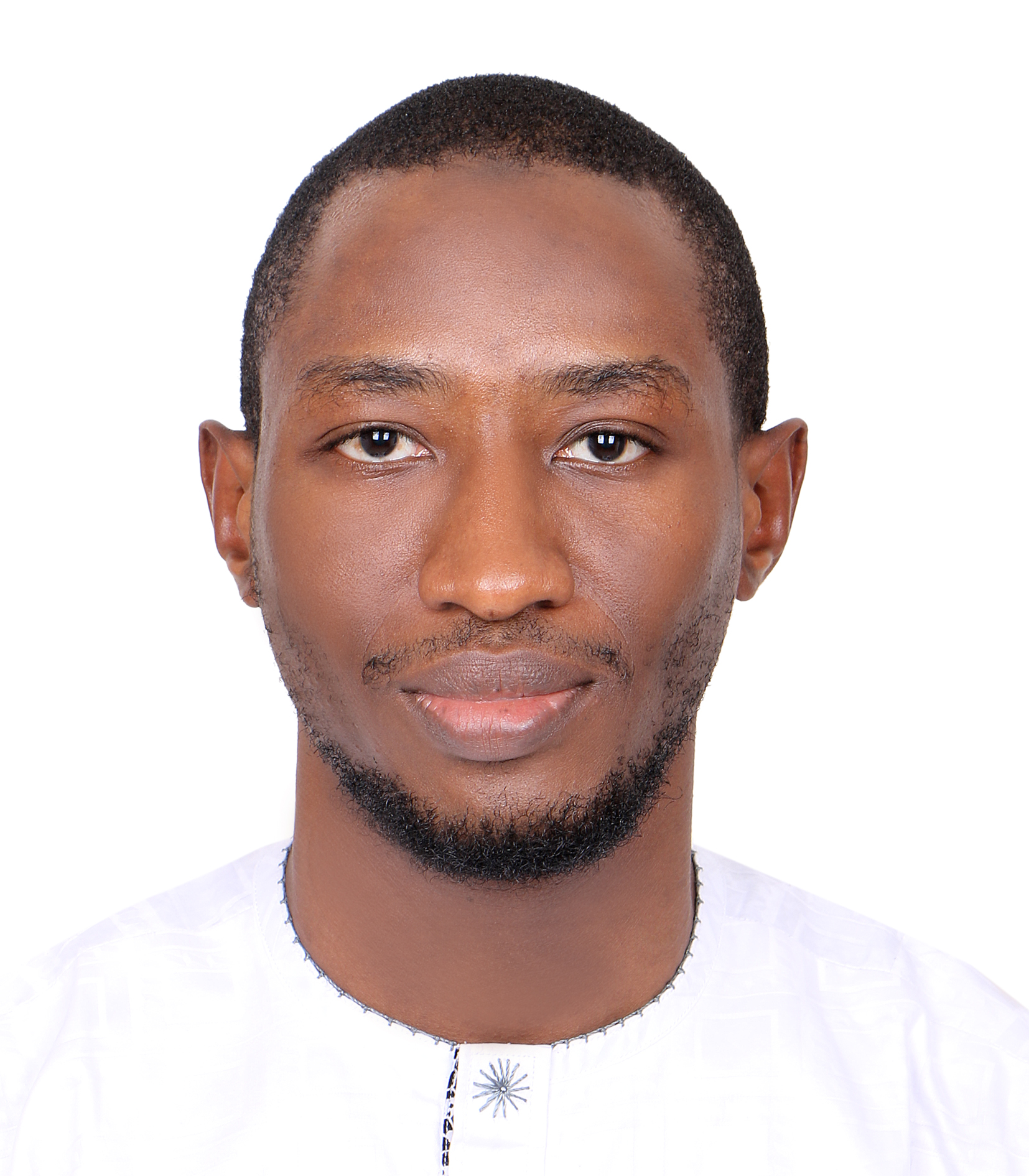}}]{Abubakar S. Ali} received the B.S. degree in electrical engineering from Bayero University Kano, Kano, Nigeria, in 2014, the M.S. degree in communications and signal processing from University of Leeds, Leeds, UK, in 2015, and the Ph.D. degree in electrical and computer engineering at Khalifa University, Abu Dhabi, United Arab Emirates, in 2023. His research interests include low-power wireless communications, machine learning, optimization for communications and networking, and security in intelligent communication systems.
\end{IEEEbiography}

\begin{IEEEbiography}[{\includegraphics[width=1in,height=1.25in, clip, keepaspectratio]{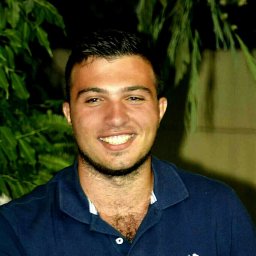}}]{Dimitrios Manias} (Member, IEEE) is a PhD student in the Optimized Computing and Communications (OC2) Lab at Western University, Ontario, Canada. He received his B.E.Sc. degree in electrical and computer engineering (2018) and his M.E.Sc. degree in electrical and computer engineering with a collaborative specialization in artificial intelligence (2019) from Western University. His current research interests include machine learning, advanced analytics, computer networks, and next-generation communication systems. He is an active IEEE volunteer and is the Secretary of the IEEE London Section.
\end{IEEEbiography}

\begin{IEEEbiography}[{\includegraphics[width=1in,height=1.25in, clip, keepaspectratio]{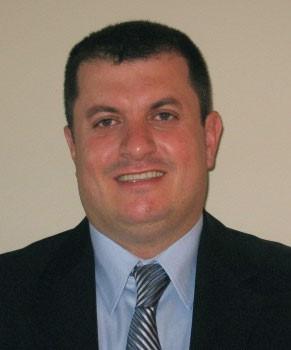}}]{Abdallah Shami} (Senior Member, IEEE) earned his Ph.D. in electrical engineering from The City University of New York in 2003. He is now a Professor at Western University, Canada, in the Electrical and Computer Engineering Department, also serving as the Acting Associate Dean (Research) and Director of the Optimized Computing and Communications Laboratory. He has held leadership roles in IEEE committees and conferences. Presently, he is an Associate Editor for publications including the IEEE Transactions on Mobile Computing and IEEE Network.
\end{IEEEbiography}

\begin{IEEEbiography}[{\includegraphics[width=1in,height=1.25in, clip, keepaspectratio]{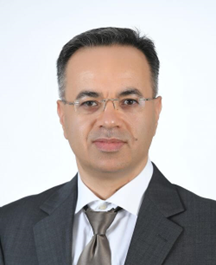}}]{Sami Muhaidat.} (Senior Member, IEEE) received the Ph.D. degree in Electrical and Computer Engineering from the University of Waterloo, Waterloo, Ontario, in 2006. He is currently a Professor at Khalifa University, and an Adjunct Professor with the Department of Systems and Computer Engineering, Carleton University, Canada. His research focuses on wireless communications, optical communications, IoT with emphasis on battery-less devices, and machine learning. 
\end{IEEEbiography}

\end{document}